\documentclass[twocolumn, 10 pt, conference]{ieeeconf}  % Comment this line out if you need a4paper

\IEEEoverridecommandlockouts                              % This command is only needed if 
\usepackage{easyReview}
\usepackage{graphics} % for pdf, bitmapped graphics files
\usepackage{subcaption}
\usepackage{multirow}
\usepackage{epsfig} % for postscript graphics files
\usepackage{mathptmx} % assumes new font selection scheme installed
\usepackage{times} % assumes new font selection scheme installed
\usepackage{amsmath} % assumes amsmath package installed
\usepackage{amssymb}  % assumes amsmath package installed
\usepackage{amsfonts,bm}
\usepackage{stfloats}
\usepackage{stackengine}
\usepackage{algorithm}
\usepackage{algorithmic}
\title{\LARGE \bf
Histo-Planner: A Real-time Local Planner for MAVs Teleoperation based on Histogram of Obstacle Distribution
}

\author{Ze Wang, Zhenyu Gao, Jingang Qu, Pascal Morin% <-this % stops a space
\thanks{*Authors are with Sorbonne Universite, CNRS, UMR 7222, Institut des Systemes Intelligents et de Robotique - ISIR, 75005 Paris, France
        { \{first.last\}@isir.upmc.fr}}%
}

\begin{document}

\maketitle
\thispagestyle{empty}
\pagestyle{empty}

% =========================================================================================
% 										ABSTRACT
% =========================================================================================
\begin{abstract}

This paper concerns real-time obstacle avoidance for micro aerial vehicles (MAVs). Motivated by teleoperation applications in cluttered environments with limited computational power, we propose a local planner that does not require the knowledge or construction of a global map of the obstacles. The proposed solution consists of a real-time trajectory planning algorithm that relies on the histogram of obstacle distribution and a planner manager that triggers different planning modes depending on obstacles location around the MAV. The proposed solution is validated, for a teleoperation application, with both simulations and indoor experiments. Benchmark comparisons based on a designed simulation platform are also provided.

\end{abstract}

% =========================================================================================
% 										INTRODUCTION
% =========================================================================================
\section{INTRODUCTION} \label{introduction}
Micro aerial vehicles (MAVs) are used in many applications, such as rescue search, forestry monitoring, infrastructure maintenance, aerial photography, etc. When the MAV operates in cluttered environments, obstacle avoidance is a major problem. Solutions to this problem are highly dependent on the type of environment, the available onboard sensors, the availability of a global map of the environment, and the available computational power. While solutions to this problem rely on both perception and planning/navigation aspects (the classical {\em sense and avoid} scenario), the present paper focuses on the navigation aspect.

Many traditional navigation methods are summarized in detail in \cite{lavalle2006}. They can be divided into three categories, namely {\it i)} Graph-based discrete path planning algorithms: A$^{*}$, RRT$^{*}$ (Rapidly-exploring random tree), PRM (Probabilistic roadmap), Visibility Roadmap, {\it ii)} Gradient-Field-based continuous path planning algorithms: Artificial Potential Field (APF) \cite{Hwang1988}, Harmonic potential field \cite{Huber2019} and Gradient-based Trajectory Optimization (GTO), etc., and {\it iii)} Goal-conscious Local obstacle avoidance algorithms: Vector Field Histogram (VFH)\cite{Borenstein1991} \cite{Ulrich1998}, Dynamic window approaches (DW) \cite{Missura2019}, Geometric Guidance (GG)\cite{Chakravarthy1998}, Streamline methods\cite{Waydo2003} etc. Local obstacle avoidance algorithms\cite{lavalle2006} are also called "path searching algorithms". 

\begin{figure}[t]
	\centering
	\includegraphics[scale=0.9]{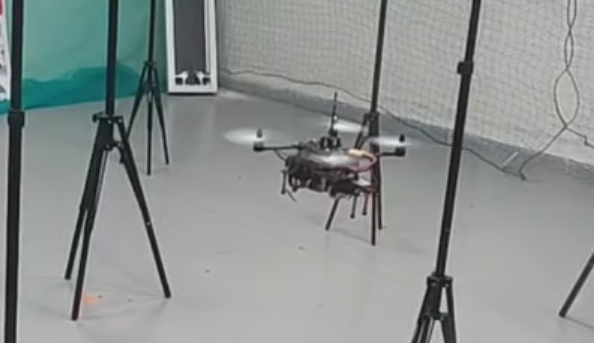}
	\caption{The custom-built MAV flight platform of with a size of 15x20x20cm and weights about 500g. The MAV is equipped with Pixhawk 4 autopilot and Khadas Vim 3. All modules run on the on-board computer.}
	\label{figure:drone}
\end{figure}

Usually navigation maps are divided into free space, occupied space and unknown space. Due to its completeness, the discrete path planning algorithm represented by A* is well suited for scenarios where the global map is known. When a global map is not available, people usually prefer frontier-based algorithms \cite{Dai2020} for space exploration tasks, which are also regarded as a local obstacle avoidance algorithm that focuses on the local obstacles. Continuous path planning algorithms can generate a trajectory with high-order continuity to be tracked by the MAV, but this algorithm often falls into local minima. Harmonic potential field, inspired by fluid mechanics, skillfully avoids local minima but introduces a large amount of calculation. However, continuous path planning methods, such as gradient-based algorithms, can be combined with frontier based methods where the latter help avoiding local minima. Thus, combining a continuous trajectory planning algorithm with a goal-conscious local obstacle avoidance algorithm can be used as a robust local trajectory planner avoiding local minima and obstacles.

Once a first collision-free trajectory has been obtained, a replanning is often performed. Recent related works \cite{gao2017}-\cite{ego2020} reveal that gradient-based trajectory optimization approaches typically formulates trajectory replanning as an optimization problem. Different methods of trajectory description have been used in these works such as polynomial curves \cite{gao2017}, \cite{Oleynikova2016}, Bézier curves \cite{Gao2018}, or  B-spline curves \cite{zhou2020}, \cite{Usenko2017}, \cite{ego2020}, \cite{Elbanhawi2015}. \cite{Usenko2017} provides that the change of an individual control point of B-splines only affects the nearest curve segments, which is important to generate a reliable trajectory. An approach that avoids the need of ESDF and only relies on the occupancy grid is proposed in \cite{ego2020}. Differently, we aim to generate local obstacle avoidance trajectories for non-globally known environments while avoiding global static occupancy grid updates and saving further computation time.

In this paper, we propose a "guided" gradient-based optimization  trajectory. Each time trajectory planning is necessary, a "guidance point" is  generated based on the local obstacle distribution, the MAV's state, and the goal location. This guidance point allows us to define an initial trajectory for the optimizer, that is helpful to avoid the local minima problem. It is generated based on a histogram of local obstacles that avoids the need to update an occupancy grid, thus saving significant computation time. Then, an optimizer is applied to generate the MAV's reference trajectory. The action of the optimizer can be seen as someone releasing an elastic rope with fixed end points from the guidance point. Additionally, a planner manager is proposed to trigger different navigation modes based on the goal point and the MAV's current location with respect to the obstacles.

The paper is organized as follows. Sect. \ref{Overview} gives an overview of the proposed navigation architecture. Sect. \ref{histo} discusses the Histogram of Obstacle Distribution generation and guidance point generation. Sect. \ref{optimization} describes the gradient-based trajectory optimization and Sect. \ref{manager} describes the planner manager. Simulation and experimental results are reported in Sect. \ref{experiment}, together with benchmark comparisons. 

% =========================================================================================
% 								Proposed Solution
% =========================================================================================
\section{Overview} \label{Overview}
\begin{figure}[t]
	\centering
	\includegraphics[scale=0.45]{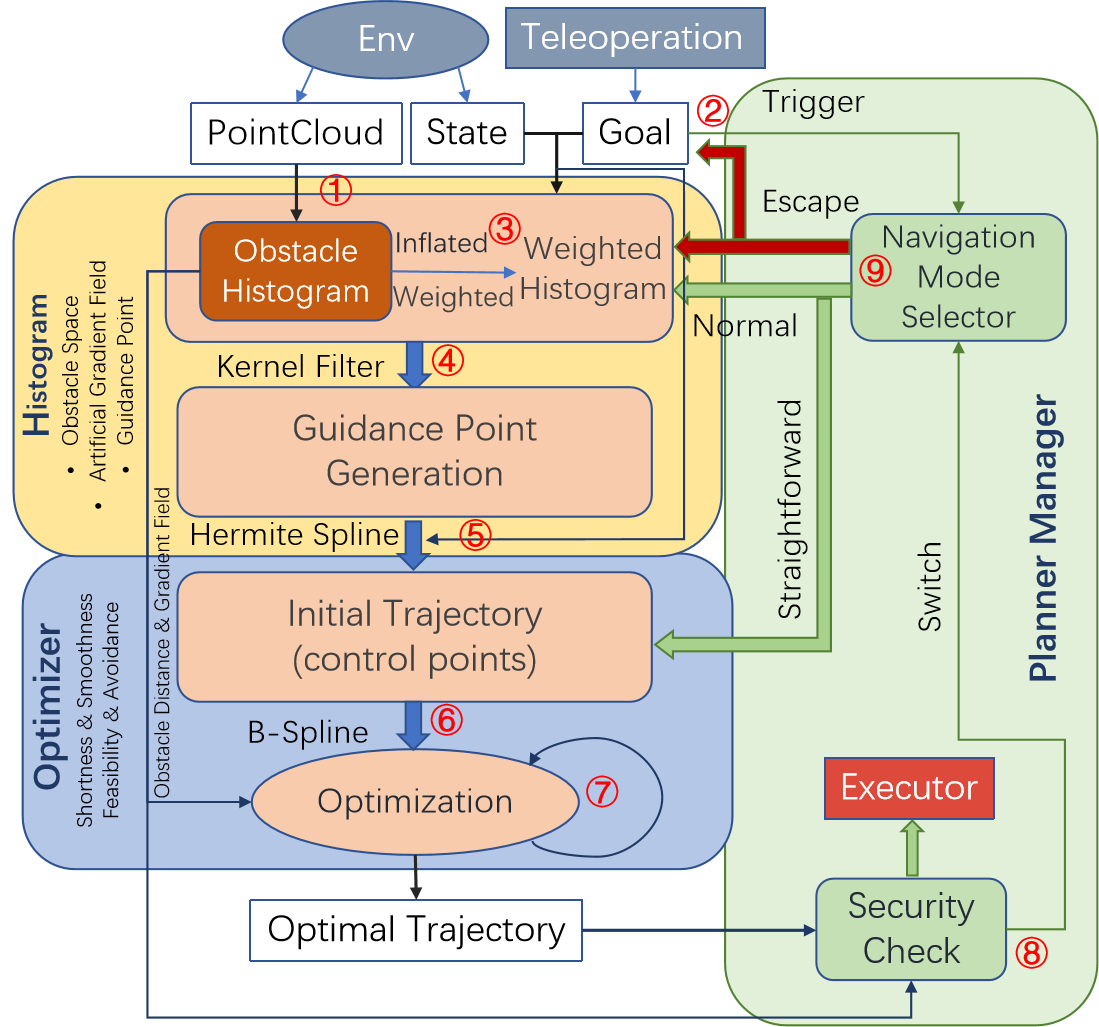}
	\caption{A global view of the navigation system architecture. }
	\label{figure:arch}
\end{figure}

Fig.\ref{figure:arch} shows the architecture of the proposed navigation system. The system is divided into three main parts: the histogram initialization and guidance point generation detailed in Sec.\ref{histo}, the optimizer detailed in Sec.\ref{optimization}, and the planner manager detailed in Sec.\ref{manager}. The navigation system performs in the order marked in the diagram.

\textbf{Input Information:} The system uses as inputs the current point clouds of the local environment, estimated MAV's states and goal point. Point clouds are typically obtained from depth camera or lidars. MAV's states contain the flight speed and position of the vehicle in local frame. The goal point can be specified by different means, depending on the information provided to the pilot. For teleoperation applications, goal point may be provided through a ground station, or even directly with a joystick.

\textbf{Histogram and Guidance point generation:} The navigation system starts with the histogram generation. The obstacle histogram is updated as a separate thread in real time from the received sensor point clouds. The histogram is used to describe the local obstacle space by a limited number of point clouds, providing obstacle distance monitoring and local gradient field generation when needed. The histogram has an image-like data structure and thus nearby obstacles have a higher resolution. Once the trajectory generation loop is activated, the obstacle histogram inflates the obstacle space according to the safety distance, and combines information of current flight states and goal location in order to generate a weighted histogram. Leveraging image processing methods, a filter kernel acting on the weighted histogram is used to find the best obstacle gap leading to the goal, thereby generating the "guidance point".

\textbf{Optimized trajectory:} Reference trajectories for the MAV are defined as  B-splines, with associated control points. The location of the control points are the variables of the optimization process. In order to initiate the optimization algorithm, first a Hermite interpolation spline passing through the guidance point is defined, and next it is transformed into a B-spline. A quadratic unconstrained optimization solver is employed here.

\textbf{Planner Manager:} The Planner Manager is responsible for safety monitoring, tracking error monitoring, and navigation state and mode switching. The reference trajectory generation loop starts with the receipt of a goal point, and a navigation mode is then triggered based on the current environment. Once a reference trajectory is generated, the security check module monitors the reference trajectory for collision risk evaluation. If a collision risk is detected, a new reference trajectory generation is triggered. An escape mode may be activated to avoid deadlocks.

% =========================================================================================
% 								SECTION 3: Histogram
% =========================================================================================
\section{Histogram and Guidance point Generation} \label{histo}

\begin{figure}[t]
	\centering
	\includegraphics[scale=0.5]{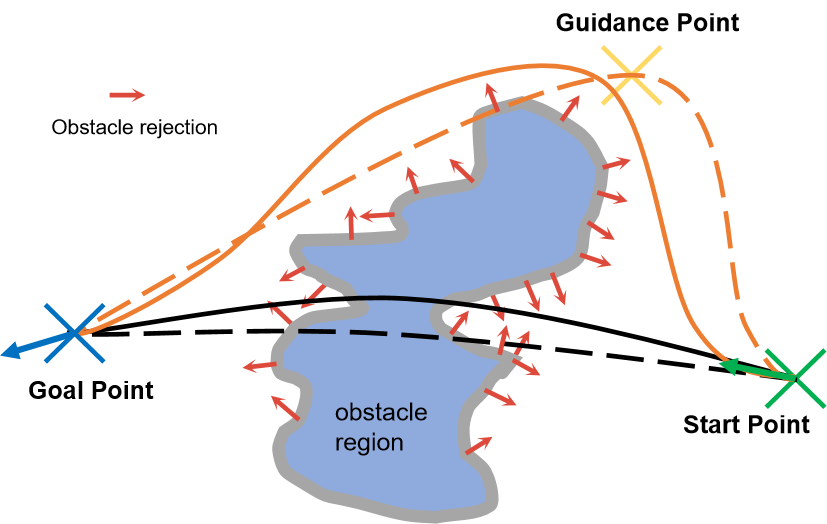}
	\caption{The dotted curve represents the initial trajectory, and the solid curve represents the optimized trajectory. The guidance point is generated from the histogram of obstacle distribution. The blue area is the obstacle area and the gray depicts its boundary.}
	\label{figure:failure}
\end{figure}

It is well known that gradient-based trajectory optimization may lead to local minima problems.
A typical example is illustrated in Fig.\ref{figure:failure}. During the planning process, the points on the trajectory should be continuously corrected to move away from the obstacles. However, if the initial trajectory falls into a corner as marked by the black dotted curve, where obstacles near the trajectory take the shape of an hourglass, obstacles will force the trajectory to get stuck there, resulting in the black solid curve. Generally, such a "corner" is difficult to avoid, especially in cluttered environments. 
In order to handle this issue, \cite{zhou2020} uses a Visibility PRM method to generate a topological pre-path in a global map. The visibility PRM approach relies on a set of points where each guard point represents a convex free space and the connection point lies on the boundary of two adjacent convex free spaces. In this work we construct a histogram of obstacle distribution to describe the local convex free space and generate a guidance point in the area of obstacles gap. Thus an initial reference trajectory passing through the guidance point is generated.

\subsection{Histogram of Obstacle Distribution}
\begin{figure}[t]
	\centering
	\includegraphics[scale=0.6]{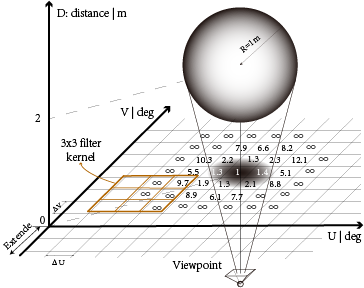}
	\caption{An obstacle appears in the field of view of the sensor. Measurements are projected onto a spherical coordinate system which is flattened into a Cartesian-like coordinate system.}
	\label{figure:VFH}
\end{figure}

Drawing on the basic idea of VFH \cite{Ulrich1998}, we design a histogram centered on the current viewpoint to describe the local space based on the sensory information. A typical spherical histogram is illustrated in Fig.\ref{figure:VFH}. An obstacle appears in the field of view of the sensor, and the point clouds' distance measurements are described in a spherical coordinate system $\Sigma_{UVD}$. The coordinate system plane $P_{UV}$ is divided into a finite number of cells of equal resolution $\Delta U \times \Delta V$, which may be much rougher than the sensor resolution, and to each cell $i=(u,v) \in P_{UV}$ is associated its
distance value $D(i)$ to the closest obstacle. As a result, such a histogram has an image-like structure, which looks like latitude and longitude lines on the world map. Thus, each cell $i$ can be viewed as a direction in the histogram frame. Obviously, $U,V,D$ are bounded, the value range of $D$ is within the distance range of the sensor, the horizontal range of $U$ is within $[-180^\circ, 180^\circ]$ and the vertical range of $V$ is within $[-90^\circ, 90^\circ]$. The obstacle histogram is updated in real time from the sensor measurements. The sensor usually has a limited field of view, and therefore its measurements are only used to update a part of the histogram. In order to cope with the dynamic environment, historical measurements within a short-term memory time window and a proximity space are used for updating the rest of the histogram. The histogram is built in the ENU 
(East-North-Up) coordinate system, with the vertical axis representing the height direction and the horizontal axis representing the horizontal circumferential direction.

\subsection{Weighted Histogram and Guidance Point Generation} \label{kernel filtering}
\begin{figure}[t]
	\centering
	\includegraphics[scale=0.2]{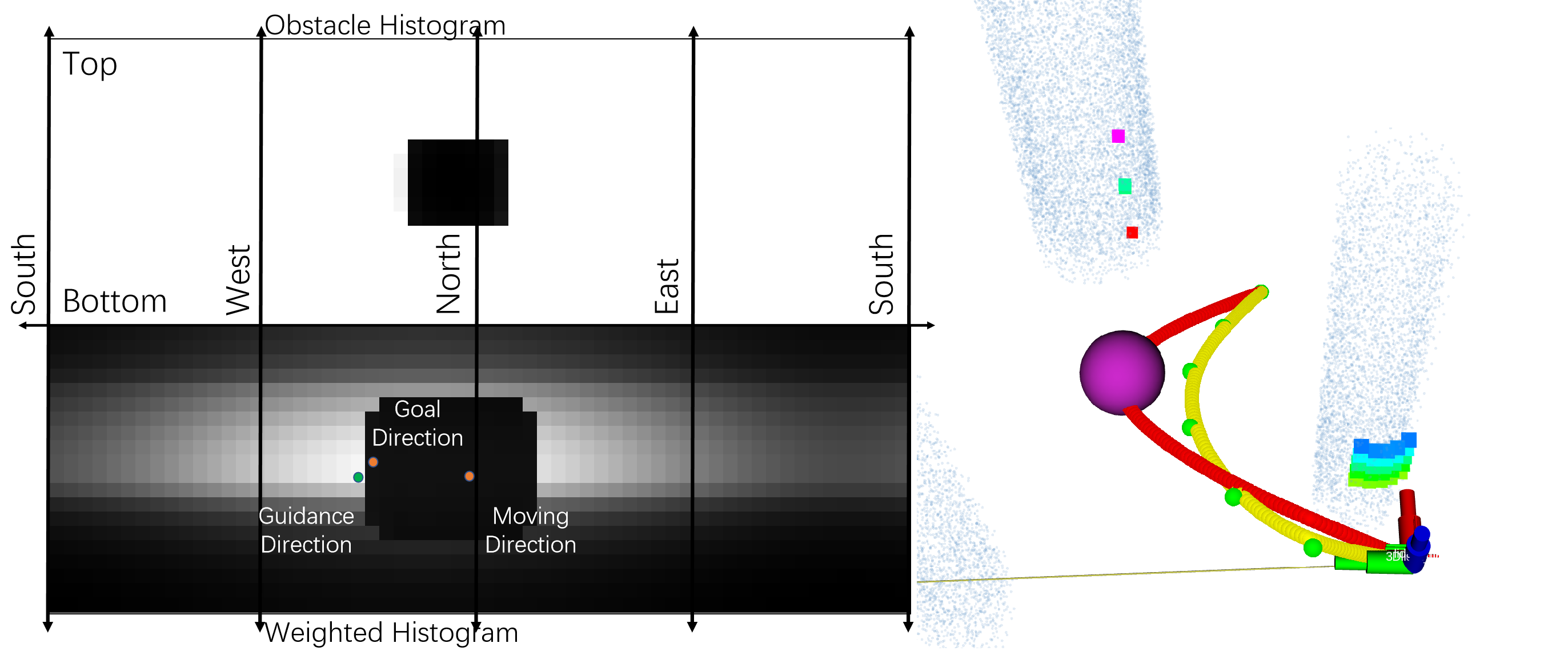}
	\caption{The right sub-figure features the vehicle's frame, guidance point (purple), initial trajectory (red), and optimal trajectory (yellow) based on the perceived environment. The obstacle histogram (upper-left sub-figure) is updated in real time based on the sensor information (color points), with the vertical axis representing the height direction and the horizontal axis representing the horizontal circumferential direction. The weighted histogram (lower-left sub-figure) is generated only if guidance point generation is required. The brightest spot represents the direction of the guidance point.}
	\label{figure:Histogram}
\end{figure}

In image processing, people usually design and use some special kernels (e.g. Edge detection, Pooling, etc.) to extract features. In this paper, we adopt the idea of kernel to extract the obstacle boundary information in the histogram space and then obtain a "guidance point". An illustative example is shown in Fig.\ref{figure:Histogram}, which features a 3D view of obstacles and MAV's trajectory (right), the histogram of obstacle distribution (top-left), and the weighted histogram (bottom-left). The weighted histogram is generated only if guidance point generation is required. Since the histogram is spherical and is therefore closed around the sides, there is no reduction in the histogram size during the kernel convolution and thus no patch is needed. 

The weighted histogram is generated in two steps. Firstly, an inflated obstacle histogram is generated from
the original obstacle histogram based on a requirement of safety distance to the obstacles. $D_{obs}(i)$ is the value of cell $i$ in this inflated obstacle histogram, which
corresponds to the distance between the MAV and the surface of the nearest inflated obstacle in the direction of cell $i$. Secondly, a weight is applied to each cell's value of the inflated histogram based on the current velocity and goal location. The value of a cell $i$ in the weighted histogram is defined as follows:
\begin{equation}
		Value(i) = (\alpha_1 V_v^{g}(i)\cdot V_u^{g}(i) + \alpha_2 V_v^{c}(i)\cdot V_u^{c}(i)) * D_{obs}(i) 
\end{equation}
where $V_v^{g}(i)$ and $V_u^{g}(i)$ (resp. $V_v^{c}(i)$ and $V_u^{c}(i)$) correspond respectively to the weights of cell $i$ in the $v$ and $u$ dimensions of the histogram considering only the goal (resp. considering the current velocity direction). $\alpha_1$ and $\alpha_2$ are non-negative trade-off weights 
with $\alpha_2$ automatically reset to zero if the MAV's velocity is too low. 
Due to similarity, the weights $V_\star^g$ and $V_\star^v$ for $\star \in \{u,v\}$ have the same mathematical expression in the spherical coordinate system, which is:
\begin{equation}
	V_\star^x(i)=(1-\_min\_)(\frac{cos((j-i)\Delta_x)+1}{2})^{\_pow\_}+\_min\_
\end{equation}
where $j$ is the index of the cell that indicates the goal direction (if $x= g$) or the current velocity direction (if $x=v$), $\Delta_x= \Delta U$ if $\star=u$ 
and  $\Delta_x= \Delta V$ if $\star=v$, and $\_min\_$, $\_pow\_$ are constant parameters. The smaller the value of $\_min\_$ or the larger the value of $\_pow\_$, the more the importance of the desired direction is emphasized. 

The guidance point is defined from the weighted histogram in two steps. The first step consists in selecting the direction of the guidance point in the current local frame. The second step consists in selecting the distance of the guidance point in this direction. The guidance point direction is the direction 
$Dir^{*}$ associated to the cell
\begin{equation}
		i^{*}= arg\max_{i}\{Avg(kernel_i) + Min(kernel_i)\} 
\end{equation}
where $Avg(kernel_i)$ (resp. $Min(kernel_i)$) is the mean value (resp. minimum value) of all cells of the weighted histogram selected by the kernel centered at cell $i$. Referring to Fig. \ref{figure:Histogram}, this direction corresponds to the brightest area of the weighted histogram.
Finally, we set
\begin{equation}
		D^{*}= min(Avg(kernel_{i^{*}}),\alpha D)
\end{equation}
with $\alpha \leq 1$ a constant positive parameter and $D$ the current distance to the goal point, and we define the guidance point as the point located at distance $D^{*}$ along the direction $Dir^{*}$.

% =========================================================================================
% 								SECTION 4: GTO
% =========================================================================================
\section{Gradient-based Optimized Trajectory Generation} \label{optimization}
B-spline are used to describe reference trajectories. A B-spline is a piecewise polynomial parametric curve \cite{Usenko2017}\cite{De Boor1972}\cite{Hartmut2002} uniquely determined by its degree $p$, a set of $N + 1$ $n$-dimensional control points $\{ P_0, P_1, \ldots , P_N \} \in \mathbb{R}^n$ and a knot vector. Given the guidance point determined in the previous section, a Hermite interpolation spline passing through this point acts as the initial trajectory. A Hermite spline \cite{Kreyszig2007} is a parametric interpolation curve uniquely determined by the endpoint constraints, such as position, tangent vector, etc.. A sequence of way-points is obtained by splitting the Hermite spline with constant time intervals. Constructing a least squares regression, we can determine a set of uniform B-spline control points based on the way-points, thus transforming the Hermite spline into a B-spline. These control points are the initial optimized variables. The optimization formulation is built below, and an example is shown in Fig.\ref{figure:Histogram}.

A $3^{rd}$-degree uniform B-spline trajectory is uniquely determined by a set of $N+1$ $n$-dimensional control points and has $(N+1)n$ degrees of freedom (DoF). The head and tail constraints on the trajectory including the position, velocity and acceleration play a role of boundary conditions, and the trajectory therefore has $(N+1-2\times3)n$ DoF at least. The optimized variables and cost function are defined as:
\begin{equation}
	\begin{aligned}
		\min_x\ f_{total}(x)=&f_{st}(x) + f_{fea}(x) + f_{smth}(x) + f_{ca}(x) \\
		\ subject\ to:\ & x \in \mathbb{R}^{(N+1)n} \ ,\ x_{0}\ ,\ x_{N}\\
		& |\omega_z| < W_{max} \ ,\ |v| < V_{max} \ ,\ |a| < A_{max} 
	\end{aligned}
\end{equation}
where $f_{st}$ is the cost of strength for anti-tensile effect and anti-bending effect, $f_{fea}$ is the cost for the velocity/acceleration/yaw-rate feasibility, $f_{smth}$ is the smoothness cost, and $f_{ca}$ is the collision avoidance cost. These cost functions are given different weights depending on the user's needs. The optimization variable $x \in \mathbb{R}^{(N+1)n}$ denotes those $N+1$ control points, which may not include all of the head $3$ control points, the tail $3$ control points.  

For the cost of strength, we design an "anti-tensile effect" to reduce the length of trajectory, and an "anti-bending effect" to prevent excessive curvature:
\begin{equation}
	\begin{aligned}
		f_{st }&=\sum_{i=0}^{N-2}||P_{i+1}-P_{i}||^2 + \sum_{i=0}^{N-2}||accel\_nor_{i}||^2
	\end{aligned}
\end{equation}
where $accel\_nor$ denotes the normal acceleration of the trajectory, defined as the cross product of the acceleration vector and the normalized velocity vector. For the smoothness cost, we use the third-order derivative of B-spline as a metric \cite{zhou2020}:
\begin{equation}
	\begin{aligned}
		f_{smth}&= \sum_{i=0}^{N-3}||P_{i+3}-3P_{i+2}+3P_{i+1}-P_{i}||^2
	\end{aligned}
\end{equation}
According to the convex hull property of B-spline, kinematic feasibility constraints are used to limit the velocity, acceleration, and yaw rate when they exceed the maximum limit. The yaw rate is obtained by dividing the normal acceleration of the trajectory by the velocity.

The collision cost is formulated as repulsive force on $Q_k$ point, referring to the artificial potential field \cite{Hwang1988}:
\begin{equation}
	f_{ca}= \sum_{i=0}^{N-2} Rep(dis(Q_i-Obs))
\end{equation}
where $Q_i=\frac{1}{6}(P_i+4P_{i+1}+P_{i+2})$ is the point on the B-spline determined by the sequential control points $P_i, P_{i+1}, P_{i+2}$, $Obs$ is the closest obstacle in the obstacle histogram, $dis(\cdot)$ is the Euclidean distance function, and $Rep(\cdot)$ is a differentiable repulsive force potential 
field \cite{Hwang1988} with a minimum distance $d_{min}$ and a maximum effective range $d_{max}$:
\begin{equation}
	Rep(d)=
	\begin{cases}
		a-10d&,d \in [0,d_{min}] \\
		5(d_{max}-d)-5b\cdot sin(\frac{d-d_{min}}{b})&,d \in [d_{min}, d_{max}] \\
	\end{cases}
\end{equation}
with $a=5(d_{min}+d_{max})$, and $b=\frac{d_{max}-d_{min}}{\pi}$. For the smoothness of the descent gradient direction, we calculated the combined force of all repulsive forces from obstacles within the maximum effective range $d_{max}$, and then rotate the gradient direction of the loss function $f_{ca}$ so that it coincides with the direction of the combined force.

% =========================================================================================
% 								SECTION 5: Navigation Mode Selector
% =========================================================================================
\section{Planner Manager} \label{manager}
Since the navigation system does not rely on a global map, new information obtained during the flight (point clouds, tracking errors) may lead to replanning decisions. This is the role of the planner manager. More precisely, the planner manager is responsible for collision risk monitoring (security check), tracking error monitoring, navigator state switching and navigator mode switching. Navigation states include hovering and waiting for a new goal point, reference trajectory generation, trajectory tracking, and replanning. A new goal point triggers a navigation state which enables the reference trajectory generation (as described in Sec. \ref{Overview}), followed by trajectory tracking. If a FOV-limited sensor is used, the vehicle will be required to align its heading with the trajectory during the flight state transition from hovering to tracking. During trajectory tracking, the planner will perform reference trajectory safety check and tracking error check. The safety distance of reference trajectory considers parameters "forbidden range" and "maximum tracking error". Once the trajectory is at risk of collision, the navigation state is switched. Depending on the length of time from the current moment to the moment of the collision, from short to long, the navigation state switches to stop and wait for a goal, stop and regenerate a reference trajectory, and re-plan, respectively. Once the goal point is at risk of collision, the planner regenerates one nearby and then switches the navigation state to replanning. The replanning navigation state will be triggered when the tracking error is too large.

The navigation environment is often complex. In some special scenarios, multiple consecutive generated trajectories detected by safety monitoring as a risk of collision will trigger escape mode, which allows the vehicle to track only the guidance point and thus track the goal indirectly. In addition, there is also Normal mode and Straightforward mode. Normal mode performs the entire process written in Sec.\ref{Overview}. Straightforward mode skips the guidance point generation phase. More precisely, when the goal point is located in the current visible space a straight line is the shortest trajectory. Considering the current motion state and the surrounding obstacle distribution, it is still necessary to generate a smooth continuous collision-free trajectory through optimization but the guidance point generation can be skipped to directly initialize a Hermite interpolation spline trajectory considering only the position and velocity constrains at the start point and goal point.

% =========================================================================================
% 								SECTION 6: Experimental Results
% =========================================================================================
\section{Experimental Results} \label{experiment}

\subsection{Implementation Details}
The planning framework is summarized in Sec.\ref{Overview}. We set the degree of B-spline $p$ as 3. The uniform B-spline interval time is 0.5 seconds. The maximum speed, acceleration, yaw rate and sensor detection range can be set depending on the environment. Smaller values are recommended for dense environments. The parameter "forbidden range" is set as the radius of the vehicle, and the parameter "maximum tracking error" depends on the accuracy of the controller. The safety distance defined from them are used for safety checking and trajectory optimization. Three types of histogram are available: Polar, Spherical, and Cylindrical. The size of the histogram is an important parameter. For spherical histogram, $60\times 20$ is enough for most cases and recommended. There are also some other functional parameters, such as, those concerning the use of joystick to input goal point, etc.. The trajectory optimization relies on a nonlinear optimization solver NLopt\footnote[1]{https://nlopt.readthedocs.io}, which is also used in Fast-planner \cite{gao2017}.

Leveraging the PX4 autopilot system\footnote[2]{https://github.com/PX4/PX4-Autopilot} and the Gazebo physical simulator\footnote[3]{https://gazebosim.org/home}, we built our simulation platform with reference to Prometheus\footnote[4]{https://github.com/amov-lab/Prometheus}. Gazebo provides the physical environment including position, sensor measurements, obstacle space, etc. The PX4 autopilot uses a 4-loop cascade PID controller for tracking trajectory. In addition, a random obstacle generation ros-node provides a random obstacle space. The open-source code of all modules\footnote[5]{https://github.com/SyRoCo-ISIR/DroneSys} are released as ROS packages. Readers can easily replicate all the presented results. The simulation experiments are performed on a laptop with an AMD 5800HS and Ubuntu20 OS.

\subsection{Simulation Experiment}
We simulate navigation missions in the gazebo physical environment with different types of sensors. The results are available in videos\footnote[6]{https://github.com/SyRoCo-ISIR/histo-planner}. 2D Lidar has a limited view and the navigation is restricted at the fixed height. Depth camera has a wider vertical field of view, and it can cope with general scenarios. The navigator is designed with heading tracking option to keep the sensor heading toward the position at time $t_f$ forward along the trajectory. Multi-line LiDAR has a panoramic view, does therefore not require heading tracking and is easier to navigate. The joystick is used to generate goal points. The outputs are scales within $[-1,1]$ multiplied by the parameter "maximum distance" to yield goal points. We embed a magnetometer on the joystick, so that the joystick outputs are expressed in different frames depending on the selection, including vehicle's frame (First-Person View), operator's frame (Third-Person View), ENU frame.

%\begin{figure}[t]
%	\centering
%	\includegraphics[scale=0.35]{sim.png}
%	\caption{Figures from top to bottom show simulation using depth camera, multi-line Lidar, and 2D-Lidar. The red and yellow curves are the initial trajectory and the optimized trajectory respectively. The long red curve is the flight trace. The purple point is the guidance point, the red point is the nearest obstacle point to the drone, and the green point is the goal point generated from joystick. The two frames represent the current drone state and the current reference respectively.}
%	\label{figure:sim}
%\end{figure}

\subsection{Planners Comparison}
\begin{figure}[t]
	\centering
	\includegraphics[scale=0.3]{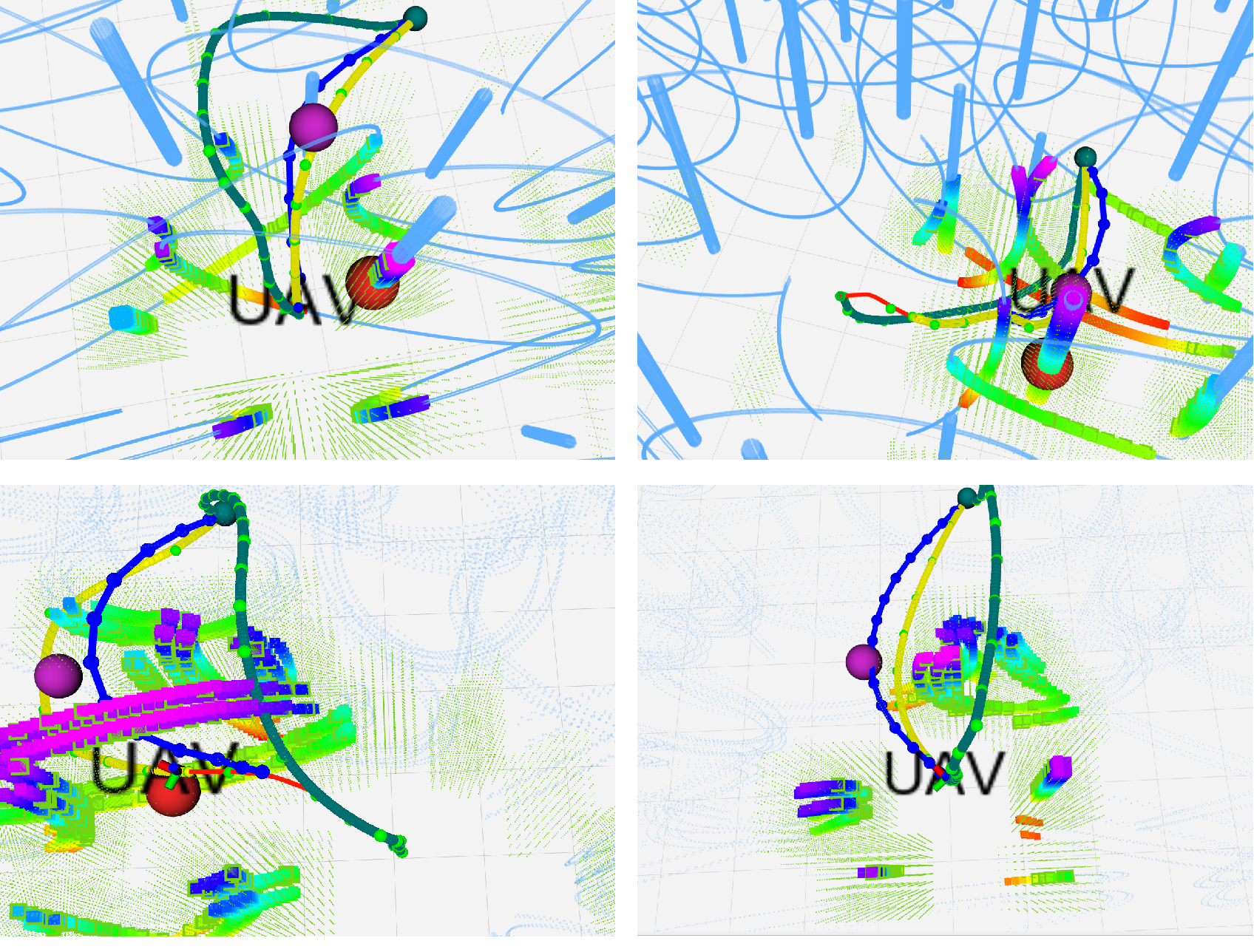}
	\caption{Different trajectory planned by the three planners for the same mission in a random obstacle environment, including 100 columns and 100 circles in a 20x20x5m space: the yellow curve (our planner), the blue curve (Ego-planner), the dark green curve (Fast-planner). The light blue point clouds are obstacles, the colored point clouds are perceived by our planner, and the green point clouds are inflated by Ego-planner. Different parameters are set as: upper-left: sensing range 2m, resolution 0.01m; upper-right: 2m, 0.05m; lower-left: 3m, 0.1m; lower-right: 2m, 0.1m.}
	\label{figure:comp}
\end{figure}

We compare our solution with two state-of-the-art planners, shown in Fig.\ref{figure:comp}. Fast-Planner \cite{gao2017} builds a 3D occupancy grid to evaluate the distance to obstacle, and then builds an ESDF to generate the gradient field. Ego-Planner \cite{ego2020} evolves from Fast-Planner and eliminates the need for ESDF. Occupancy grid has a limited range, so the planner will not work when goal point falls outside the global map. Our planner can navigate in an infinite space by not relying on it. Each planner runs dozens of times in an unknown space with different parameter settings shown in Fig.\ref{figure:comp}. The density of obstacles is set as a uniform random distribution of 100 columns and 100 circles of different sizes in a 20x20x5m space. The results show that generally our planner generates similar trajectory as Ego-planner, and that both planners outperform Fast-planner. In addition, Ego-planner generates a cubic inflated point clouds, which compresses the navigable range, especially in scenarios with narrow space, because the edges of the cube should not have been considered occupied, while our planner does not have the issue.

\begin{table}[]
	\centering
	\resizebox{0.5\textwidth}{!}{%
		\begin{tabular}{|l|l|l|ll|ll|ll|}
			\hline
			\multirow{2}{*}{Res} & \multirow{2}{*}{Range} & \multirow{2}{*}{PCL Num} & \multicolumn{2}{l|}{Ego-planner} & \multicolumn{2}{l|}{Fast-planner} & \multicolumn{2}{l|}{Histo-planner} \\ \cline{4-9} 
			&  &  & \multicolumn{1}{l|}{grid} & plan & \multicolumn{1}{l|}{esdf} & plan & \multicolumn{1}{l|}{histo} & plan \\ \hline
			
			0.01& 2& 20590& \multicolumn{1}{l|}{38.7} & \textbf{0.86} & \multicolumn{1}{l|}{250} & 12 & \multicolumn{1}{l|}{\textbf{1.4}} & 2.78 \\ \hline
			0.01& 2& 56147& \multicolumn{1}{l|}{165.2} &  & \multicolumn{1}{l|}{677.6} &  & \multicolumn{1}{l|}{\textbf{2.9}} &  \\ \hline
			0.05& 2& 2401& \multicolumn{1}{l|}{4.7} & \textbf{0.92} & \multicolumn{1}{l|}{15.6} & 3.82 & \multicolumn{1}{l|}{\textbf{0.3}} & 2.01 \\ \hline
			0.05& 2& 5300& \multicolumn{1}{l|}{13.5} &  & \multicolumn{1}{l|}{56.2} &  & \multicolumn{1}{l|}{\textbf{0.5}} &  \\ \hline
			0.1& 2& 1285& \multicolumn{1}{l|}{3.1} & \textbf{0.34} & \multicolumn{1}{l|}{12.1} & failed & \multicolumn{1}{l|}{\textbf{0.2}} & 2.66 \\ \hline
			0.1& 2& 2564& \multicolumn{1}{l|}{4.2} &  & \multicolumn{1}{l|}{20.1} &  & \multicolumn{1}{l|}{\textbf{0.3}} &  \\ \hline
			0.1& 3& 5310& \multicolumn{1}{l|}{12.4} & \textbf{1.22} & \multicolumn{1}{l|}{54.8} & 43 & \multicolumn{1}{l|}{\textbf{0.5}} & 3.43 \\ \hline
			0.1& 3& 6185& \multicolumn{1}{l|}{12.6} &  & \multicolumn{1}{l|}{60.4} &  & \multicolumn{1}{l|}{\textbf{0.6}} &  \\ \hline
			meter& meter& number & \multicolumn{1}{l|}{ms} & ms & \multicolumn{1}{l|}{ms} & ms & \multicolumn{1}{l|}{ms} & ms \\ \hline
		\end{tabular}%
	}
	\caption{Time consumed by different planners for the same navigation mission with different parameters of sensor sensing range and resolution.}
	\label{table:time}
\end{table}

Time consumed by the three planners for the same navigation mission in different scenarios is collected in Table.\ref{table:time}. The sensor sensing range and point clouds resolution determine the number of local point clouds published in real time. While the vehicle is hovering, the number of received point clouds is almost invariant. We record then the mean value of the number of point clouds received each time over a period of time in "PCL Num" column, and the mean value of the time that each planner updates the local map in the first column of each (grid, esdf, and histo respectively). Multiple navigation missions are triggered at different locations, each planner provides its solution, and the median value of the time spent on planning is recorded in the  "plan" column. The variance of time spent on updating local map is small, while that on planning changes with the mission. From the statistics, compared to our planner, others' map update time increases significantly with the number of point clouds. Although the planning time is very variable, it is independent of the number of point clouds and is related to the length of the trajectory. Ego-planner plans quickly, while our solutions respond more quickly. Illustrations are provided in videos\footnote[1]{https://github.com/SyRoCo-ISIR/histo-planner}.

\subsection{Real-world Experiment}
\begin{figure}[t]
	\centering
	\includegraphics[scale=0.4]{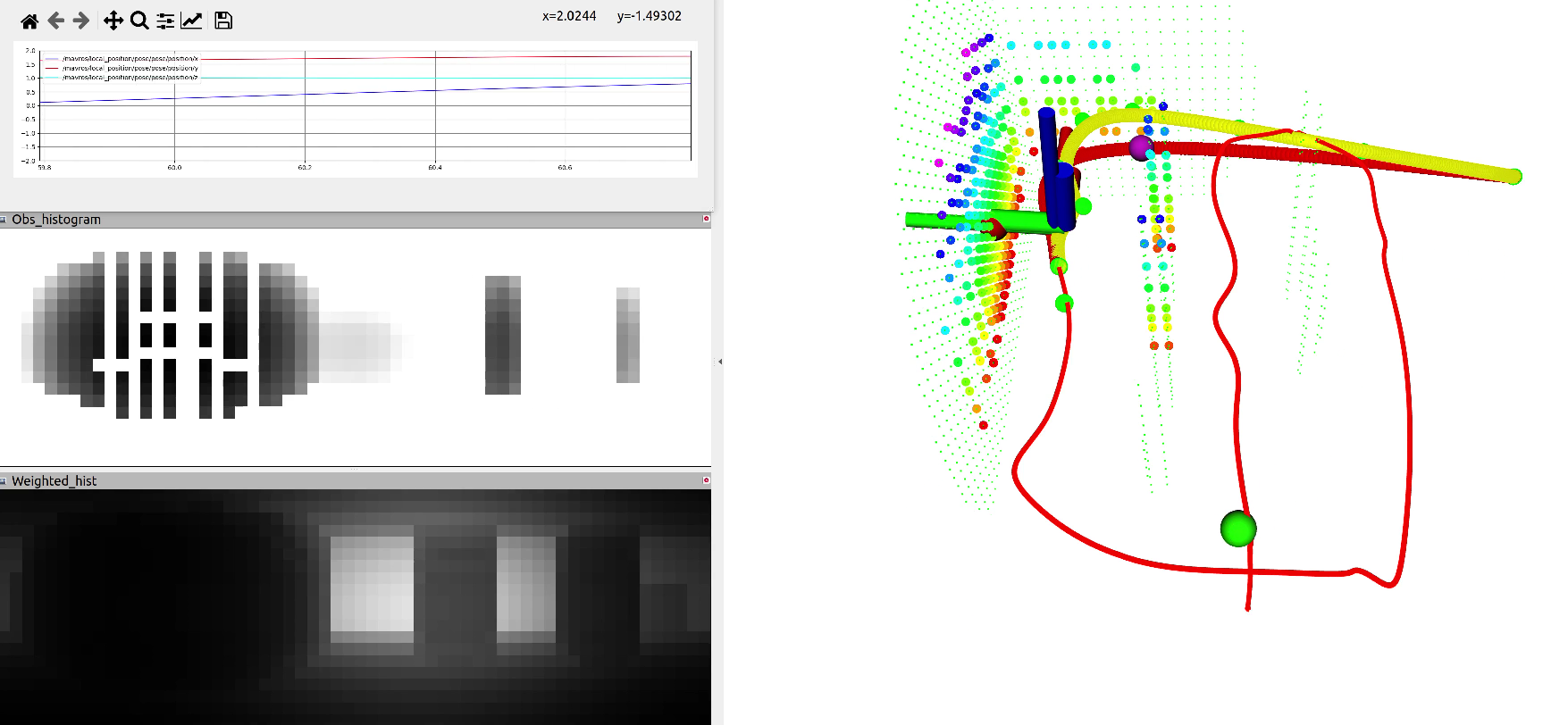}
	\caption{A screenshot showing the trajectory planning process in an indoor obstacle environment.}
	\label{figure:nav}
\end{figure}

We performed experiment in a flight arena equipped with a MoCap, used as the localization system. We pre-determine the obstacles locations (1m spacing) and have a standalone ros node to publish virtual obstacle point clouds (0.1m diameter), where obstacles correspond to tripods as shown on Fig.\ref{figure:drone}. The flight is restricted to a fence space of 4x4x2m, which is smaller than the actual MoCap space. A screenshot of the experiment video$^1$ shown in Fig.\ref{figure:nav}, the obstacle histogram shows the current distribution of obstacles in real time and the weighted histogram shows the best guidance direction. Safety parameters are set so as to maintain a safety distance not less than 0.3m from the surface of the obstacles. In this environment the vehicle basically keeps the flight speed at 1m/s, and the maximum speed can be 1.56m/s.

% =========================================================================================
% 								CONCLUSION & FUTURE WORK 
% =========================================================================================
\section{Conclusions \& Future Work}\label{conclusion}
We have proposed a novel local planner based on an image-like histogram of obstacle distribution. The planner does not rely on an occupancy grid map, which allows it to be computationnaly efficient and not to be limited by occupancy grid space. Benefiting from the image-like structure of the histogram, a deep learning-based approach will be used to generate the initial trajectory of the reference trajectory generation module in the future.
\addtolength{\textheight}{-13cm}

\section*{Acknowledgments}
The first and second authors are funded by the CSC from the Ministry of Education of the People's Republic of China.

% =========================================================================================
% 									BIBLIOGRAPHY 
% =========================================================================================

\end{document}